\documentclass{article}
\usepackage{ijcai24}

\usepackage{times}
\usepackage{soul}
\usepackage{url}
\usepackage[hidelinks]{hyperref}
\usepackage[utf8]{inputenc}
\usepackage[small]{caption}
\usepackage{graphicx}
\usepackage{amsmath}
\usepackage{amsthm}
\usepackage{booktabs}
\usepackage{algorithm}
\usepackage{algorithmic}
\usepackage[switch]{lineno}

\usepackage{amssymb}
\usepackage{mathtools}
\usepackage{xcolor}

\newtheorem{remark}{Remark}

\title{Reinforcement Learning from Diverse Human Preferences}

\author{
Wanqi Xue$^1$
\and
Bo An$^{1,2}$\and
Shuicheng Yan$^{2}$\And
Zhongwen Xu$^{3}$\thanks{Corresponding author}\\
\affiliations
$^1$Nanyang Technological University\\
$^2$Skywork AI, Singapore\\
$^3$Tencent AI Lab\\
\emails
wanqi001@e.ntu.edu.sg,
boan@ntu.edu.sg,
shuicheng.yan@kunlun-inc.com,
zhongwenxu@tencent.com}
\begin{document}

\maketitle

\begin{abstract}
The complexity of designing reward functions has been a major obstacle to the wide application of deep reinforcement learning (RL) techniques. Describing an agent's desired behaviors and properties can be difficult, even for experts. A new paradigm called reinforcement learning from human preferences (or preference-based RL) has emerged as a promising solution, in which reward functions are learned from human preference labels among behavior trajectories. However, existing methods for preference-based RL are limited by the need for accurate oracle preference labels. This paper addresses this limitation by developing a method for learning from diverse human preferences. The key idea is to stabilize reward learning through regularization and correction in a latent space. To ensure temporal consistency, a strong constraint is imposed on the reward model that forces its latent space to be close to a non-parameterized distribution. Additionally, a confidence-based reward model ensembling method is designed to generate more stable and reliable predictions. The proposed method is tested on a variety of tasks in DMcontrol and Meta-world and has shown consistent and significant improvements over existing preference-based RL algorithms when learning from diverse feedback, paving the way for real-world applications of RL methods.

\end{abstract}

\begin{figure*}
    \centering
    \includegraphics[width=0.76\textwidth]{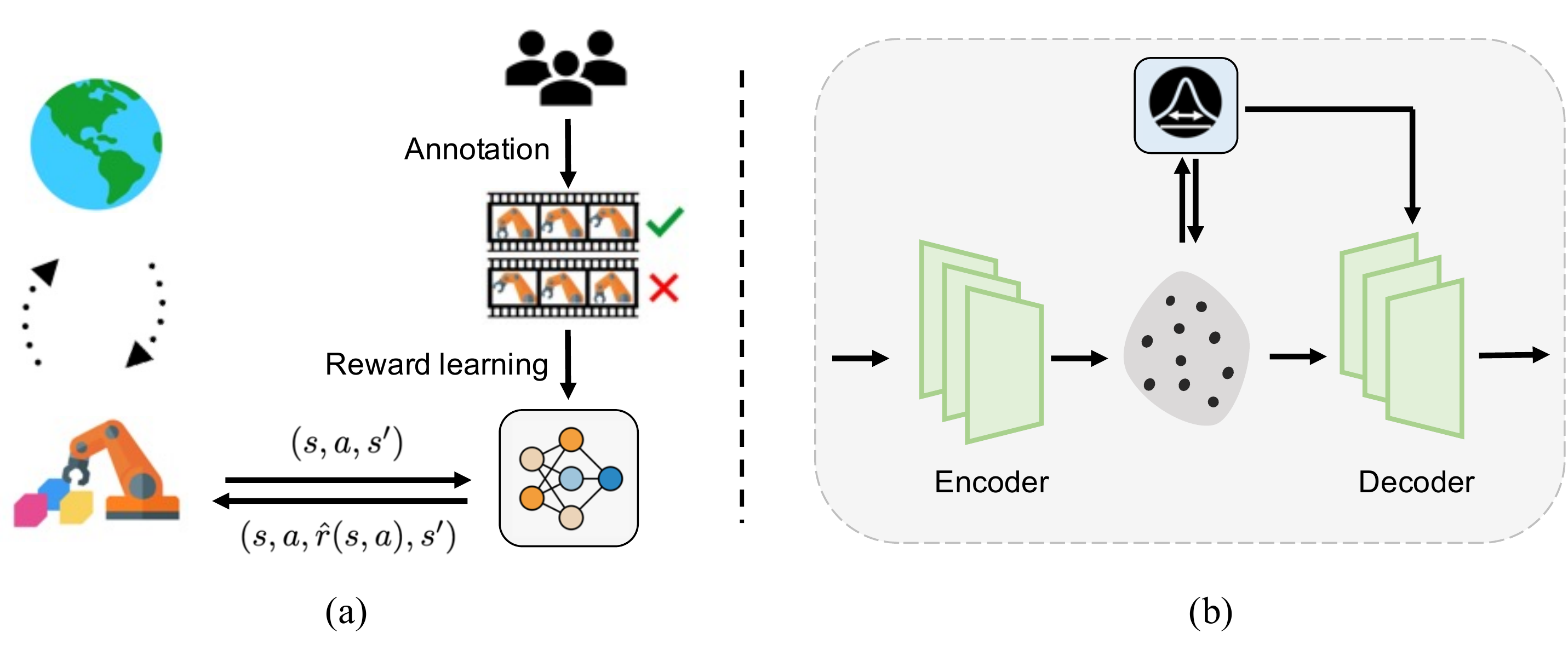}
    \caption{Illustration of our method. (a) There is a team of different annotators with bounded rationality to provide their preferences. Based on the preference data, a reward model is learned and used to provide rewards to an RL agent for policy optimization. (b) The reward models encode an input into a latent space where a strong distribution constraint is applied to address inconsistency issues. Following that, a novel reward model ensembling method is applied to the decoders to aggregate their predictions.}
    % The full procedure is summarized in Algorithm~\ref{alg}.}
    \label{fig:frame}
\end{figure*}

\section{Introduction}

Recent advances in reinforcement learning (RL) have achieved remarkable success in simulated environments such as board games~\cite{silver2016mastering,alphazero,deepstack} and video games~\cite{dqn,alphastar,wurman2022outracing}. However, the application of RL to real-world problems remains a challenge due to the lack of a suitable reward function~\cite{leike2018scalable}. On the one hand, designing a reward function to provide dense and instructive learning signals is difficult in complex real-world tasks~\cite{christiano2017deep,pmlr-v139-lee21i}. On the other hand, RL agents are likely to exploit a reward function by achieving high return in an unexpected and unintended manner~\cite{leike2018scalable,ouyang2022training}. To alleviate the problems, preference-based RL is proposed to convey human-preferred objectives to agents~\cite{christiano2017deep,stiennon2020learning}. In preference-based RL, a (human) teacher is requested to provide his/her preferences over pairs of agents' historical trajectories.  Based on human feedback, a reward model is learned and applied to provide reinforcement signals to agents (see Fig.~\ref{fig:frame}(a)). Preference-based RL provides an effective way to learn from human intentions, rather than explicitly designed rewards, and has shown its effectiveness in areas such as robotics control~\cite{pmlr-v139-lee21i} and dialogue systems~\cite{ouyang2022training}.

{\color{black}{Though promising, scaling preference-based RL to large-scale problems is still difficult because the demand for human feedback increases significantly with the complexity of problems~\cite{biyik2018batch,lee2021bpref,liang2022reward}.}
Recently, there has been a trend to replace expensive expert feedback with crowd-sourced data for scalability~\cite{gerstgrasser2022crowdplay,ouyang2022training}.
For example, in ChatGPT, a group of annotators are hired for providing affordable human feedback to RL agents. 
However, learning from crowd-sourced preferences that are inevitably unreliable, inconsistent, or even adversarial is a difficult task. The misalignment in human annotations will cause severe fluctuations in reward learning and, consequently, leading to the collapse of the policy learning.}

In this paper, we propose a simple yet effective method to help existing preference-based RL algorithms learn from inconsistent and diverse human preferences. The key idea is to stabilize the reward learning by regularizing and correcting its predictions in a latent space. Concretely, we first map the inputs of the reward model to a latent space, enabling the predicted rewards to be easily manipulated by varying them in this space. Second, to ensure temporal consistency throughout the learning process, we impose a strong constraint on the latent space by forcing it to be close to a non-parameterized distribution. This non-parameterized distribution serves as a reference point, providing a way to measure the consistency of the predicted rewards over time. Lastly, we measure the confidence of the reward model in its predictions by calculating the divergence between its latent space and the non-parameterized distribution. Based on this divergence, we design a confidence-based reward model ensembling method to generate more stable and reliable predictions. We demonstrate the effectiveness of our method on a variety of complex locomotion and robotic manipulation tasks (see Fig.~\ref{fig:demon}) from DeepMind Control Suite (DMControl)~\cite{tassa2018deepmind,tunyasuvunakool2020dm_control} and Meta-world~\cite{yu2020meta}. The results show that our method is able to effectively recover the performance of existing preference-based RL algorithms under diverse preferences in all the tasks.

\section{Preliminaries}
We consider the reinforcement learning (RL) framework which is defined as a Markov Decision Process (MDP). Formally, an MDP is defined by a tuple $\langle \mathcal{S}, \mathcal{A}, r, P, \gamma \rangle$, where $\mathcal{S}$ and $\mathcal{A}$ denote the state and action space, $r(s,a)$ is the reward function, $P(s'|s,a)$ denotes the transition dynamics, and $\gamma\in[0,1)$ is the discount factor. At each timestep $t$, the agent receives the current state $s_t\in\mathcal{S}$ from the environment and makes an action $a_t\in\mathcal{A}$ based on its policy $\pi(a_t|s_t)$. Subsequently, the environment returns a reward $r_t$ and the next state $s_{t+1}$ to the agent. RL seeks to learn a policy such that the expected cumulative return, $\mathbb{E}\left[\sum_{k=0}^{\infty}\gamma^{k}r(s_{t+k},a_{t+k})\right]$, is maximized. 

In realistic applications, designing the reward function to capture human intent is rather difficult. Preference-based RL is therefore proposed to address this issue by learning a reward function from human preferences~\cite{akrour2011preference,wilson2012bayesian,christiano2017deep,ibarz2018reward}. Specifically, there is a human teacher indicating his/her preferences over pairs of segments $(\sigma^0,\sigma^1)$, where a segment is a part of the trajectory of length $H$, i.e., $\sigma=\{(s_1,a_1),\dots,(s_H,a_H)\}$. The preferences $y$ could be $(0,1)$, $(1,0)$ and $(0.5,0.5)$, where $(0,1)$ indicates $\sigma^1$ is preferred to $\sigma^0$, i.e., $\sigma^{1} \succ \sigma^{0}$; $(1,0)$ indicates $\sigma^{0} \succ \sigma^{1}$; and $(0.5,0.5)$ implies an equally preferable case. Each feedback is stored as a triple $(\sigma^0,\sigma^1,y)$ in a preference buffer $\mathcal{D}_{p}=\{((\sigma^0,\sigma^1,y))_i\}_{i=1}^{N}$.

To learn a reward function $\hat{r}$ from the labeled preferences, similar to most prior work~\cite{ibarz2018reward,pmlr-v139-lee21i,lee2021bpref,liang2022reward,park2022surf,hejna2022few}, we define a preference predictor by following the Bradley-Terry model~\cite{bradley1952rank}:
\begin{equation}
    P_{\psi}\left[\sigma^{1} \succ \sigma^{0}\right]=\frac{\exp \left(\sum_{t=1}^H \hat{r}\left({s}_{t}^{1}, {a}_{t}^{1};\psi \right)\right)}{\sum_{i \in\{0,1\}} \exp \left(\sum_{t=1}^H \hat{r}\left({s}_{t}^{i}, {a}_{t}^{i};\psi \right)\right)}.
\end{equation}

Given the preference buffer $\mathcal{D}_{p}$, we can train the reward function $\hat{r}$ by minimizing the cross-entropy loss between the preference predictor and the actually labeled preferences:
\begin{multline}
\label{r_loss}
    \mathcal{L}_{s}=-\underset{\left(\sigma^{0}, \sigma^{1}, y\right) \sim \mathcal{D}_p}{\mathbb{E}}\Bigl[ y(0)\log P_{\psi}\left[\sigma^{0} \succ \sigma^{1}\right] \\ + y(1) \log P_{\psi}\left[\sigma^{1} \succ \sigma^{0}\right] \Bigr].
\end{multline}
where $y(0)$ and $y(1)$ are the first and second element of $y$, respectively. With the learned rewards, we can optimize a policy $\pi$ using any RL algorithm to maximize the expected return~\cite{christiano2017deep}.

\section{Preference-based Reinforcement Learning from Diverse Human Feedback}
{\color{black}Scaling preference-based RL to real-world problems necessitates a substantial amount of human feedback. However, obtaining high-quality human feedback poses a significant cost, presenting a dilemma in achieving a trade-off between the quality and quantity of human feedback. Existing works have typically focused on improving feedback-efficient, which reduce the demand for high-quality human feedback~\cite{pmlr-v139-lee21i,pmlr-v139-lee21i}. However, these methods often exhibit poor tolerance to inaccurate feedback. Low-quality human feedback could easily cause the collapse of these these approaches. In this work, we focus on addressing this issue and propose a simple yet effective method aimed at enabling RL agents to learn from diverse human feedback that possess high inconsistency. We identify that the key challenge of this problem is how to interpret temporally consistent reinforcement signals from inconsistent human feedback, i.e., how to stabilize the reward learning to provide reliable and informative guidance for policy optimization. To achieve this, we propose to implicitly manipulate the rewards by first projecting them into a latent space and then forcing them to be close to a non-parameterized distribution. Furthermore, we design a novel ensembling method to aggregate the predicted rewards. Please refer to Fig.~\ref{fig:frame} for an overview and the following sections for details.
}

\begin{algorithm}[tb]
   \caption{RL from Diverse Human Preferences}
   \label{alg}
\begin{algorithmic}[1]
   \STATE {\bfseries Input:} Strength of constraint $\phi$, number of reward models $N$, frequency of feedback session $K$
   \STATE Initialize parameters of policy $\pi(s,a)$ and the reward model $\hat{r}(s,a;\psi)$
   \STATE Initialize preferences buffer $\mathcal{D}_{p}\leftarrow \emptyset$ and
replay buffer $\mathcal{D}_{r}\leftarrow \emptyset$
   \FOR{each iteration}
   \IF{iteration $\%$ $K == 0$}
   \STATE Sample $(\sigma^0,\sigma^1)$ and query annotators for $y$
   \STATE Store preference $\mathcal{D}_{p} \leftarrow \mathcal{D}_{p} \cup \{(\sigma^0,\sigma^1,y)\}$
   \FOR{each reward model updating step}
   \STATE Sample a minibatch of preferences $(\sigma^0,\sigma^1,y)$
   \STATE Optimize the reward model (Eq.~\ref{main_l})
   \ENDFOR
   \ENDIF
   \FOR{each timestep}
   \STATE Collect $s'$ by taking action $a\sim \pi(s,a)$
   \STATE Store transition $\mathcal{D}_{r} \leftarrow \mathcal{D}_{r} \cup \{(s,a,s')\}$
   \ENDFOR
   \FOR{each policy updating step}
   \STATE Sample a minibatch of transitions $(s,a,s')$
   \STATE Measure the confidence of reward models (Eq.~\ref{confidence})
   \STATE Relabel the transitions with $\hat{r}(s,a;\psi)$ (Eq.~\ref{reward})
   \STATE Update the policy $\pi(s,a)$ on $\{(s,a,\hat{r}(s,a),s')\}$
   \ENDFOR
   \ENDFOR
\end{algorithmic}
\end{algorithm}

\subsection{Manipulating Rewards within a Latent Space}
{\color{black}Reward model will suffer from severe fluctuations if its learning signals, i.e., the human feedback, contains noise, self-contradiction, or even adversaries. Side-effects of the fluctuations will be further amplified in the policy optimization process. To deal with the problem, the predicted rewards should be corrected before they can be fed to RL agents. In general, the manipulation of rewards can be either explicitly or implicitly. For example, we can directly add or deduct a value to the predictions. However, executing such manipulations demands intricate considerations and can be as challenging as designing the rewards themselves. Therefore, we propose to implicitly manipulate the rewards within a learnable latent space. We use deep neural networks (DNNs) to project the input $(s,a)$ of the reward model into a Gaussian distribution, with the mean and the convariance matrix of the Gaussian predicted the DNNs. Formally, the DNN encoder $p(z|s,a;\psi_e)$, parameterized by $\psi_e$, is in the form of 
\begin{equation}
    p(z|s,a;\psi_e)=\mathcal{N}(z|f^{\mu}(s,a;\psi_e),f^{\Sigma}(s,a;\psi_e)),
\end{equation}
where $\mathcal{N}(\mu,\Sigma)$ denotes a Gaussian distribution with mean vector $\mu$ and covariance matrix $\Sigma$. The encoder consists of two multi-layer perceptrons whose output generates the $K$-dimensional mean $\mu$ and the $K \times K$ covariance matrix $\Sigma$, respectively. To generate rewards, there is a decoder $d(r|z;\psi_d)$, with parameters $\psi_d$, takes as input a sampled latent variable from the Gaussian and outputs a reward. Such encoder-decoder structure mainly enjoys two benefits: i) we can control the fluctuations of the rewards by adjusting the distribution of the latent space, without touching the rewards directly; ii) the confidence of the reward model to its predictions can be easily measured by calculating the divergence between different latent spaces. In the following sections, we will elaborate on how to leverage the advantages.
}

\begin{figure}
    \centering
    \includegraphics[width=0.8\linewidth]{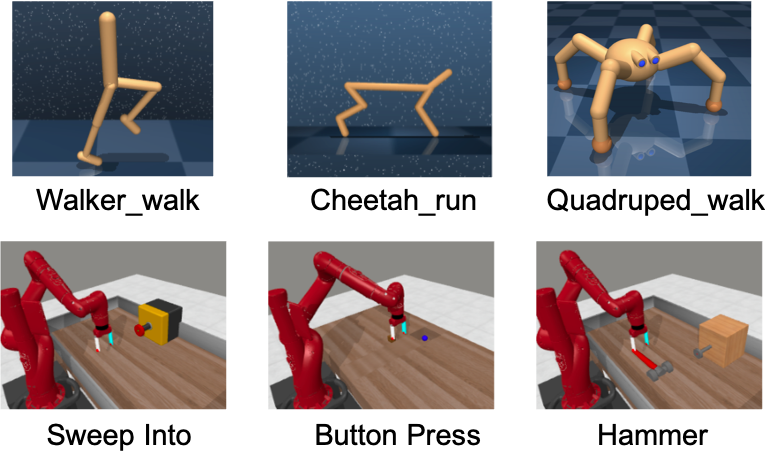}
    \vspace{-0.3cm}
    \caption{Examples for locomotion and robotic manipulation tasks.}
    \label{fig:demon}
    \vspace{-0.3cm}
\end{figure}

\begin{figure*}
    \centering
  \includegraphics[width=1\textwidth]{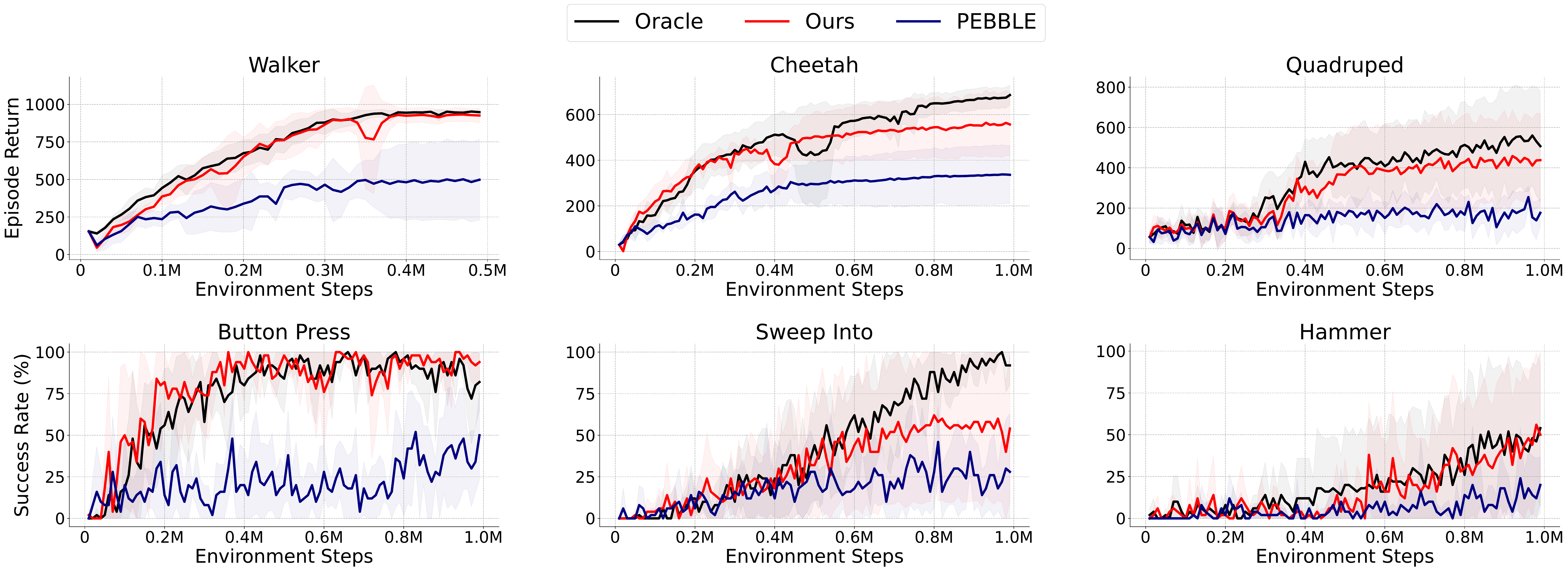}
    \caption{Learning curves on locomotion tasks (first row) and robotic manipulation tasks (second row). The locomotion tasks are measured on the ground truth episode return while the robotic manipulation tasks are measured on the success rate. The solid line and shaded regions represent the mean and standard deviation, respectively, across five runs.}
    \label{fig:main}
    \vspace{-0.3cm}
\end{figure*}

\subsection{Achieving Consistency by Imposing a Strong Constraint}
{\color{black} Our idea of achieving temporal consistency is to define a reference point for the reward model that is fixed throughout the policy optimization process. Instead of controlling the rewards directly, we set a fixed non-parameterized distribution as the reference point for the latent space of the rewards. Then, the learnable latent space is constrained to be close to the non-parameterized distribution throughout the learning process. We measure the divergence between the latent space and the non-parameterized distribution $r(z)$ by Kullback-Leibler (KL) divergence, and try to minimize:
\begin{equation}
\label{l_c}
    \mathcal{L}_c=\mathbb{E}_{(s,a)}\Bigl[ \text{KL}(p(z|s,a;\psi_e)||r(z)) \Bigr].
\end{equation}
Since $r(z)$ is non-parameterized, the optimization of the encoder will be guided toward a fixed direction throughout the learning process. With the addition of learning signals provided by human feedback, the overall loss function is 
\begin{equation}
\label{main_l}
    \mathcal{L}=\phi*\mathcal{L}_c+\mathcal{L}_s,
\end{equation}
where $\phi$ is a parameter controlling the strength of the constraint. Conventionally, $\phi$ is set as a small value to confirm that the supervised signals will not be dominated. However, counterintuitively, we found that a large $\phi$ is crucial for the expected performance. Larger $\phi$ will lead to a stronger constraint on the latent space. As a result, the latent space for different inputs will be similar, and the predicted rewards will be controlled into a smaller range. Considering that it is relative values of rewards, not absolute values, that affect a policy, a reward model with a smaller value range can lead to more accurate, stable, and effective outcomes, and therefore benefit the learning process. 
}

\begin{remark}
Controlling the divergence between the latent space and the non-parameterized distribution (Eq.~\ref{l_c}) is equivalent to minimizing the mutual information between $(S,A)$ and $Z$, which leads to a concise representation of the input.
\end{remark}

\textit{Proof.} To simplify the notation, we let variable $X$ denote the input pairs $(S,A)$. Then by the definition of KL-divergence:
\begin{equation}
    \begin{aligned}
        \mathcal{L}_c&=\iint dx\ dz \ p(x)\Big[ p(z|x) \ \log\frac{p(z|x)}{r(z)}\Big]\\
        &= \iint dx\ dz \ p(x,z)\log p(z|x)  - \int dz \ p(z)\log r(z).
    \end{aligned}
\end{equation}

Since $\text{KL}(p(z)|r(z))\geq0$, we have $\int dz \ p(z)\log p(z) \geq \int dz \ p(z)\log r(z)$. As a result,
\begin{equation}
\label{proof}
\begin{aligned}
    \mathcal{L}_c&\geq
        \iint dx\ dz \ p(x,z)\log p(z|x)  - \int dz \ p(z)\log p(z) \\
        &= I(Z,X).
\end{aligned}
\end{equation}
Eq.~\ref{proof} shows that minimizing $\mathcal{L}_c$ is equivalent to minimizing an upper bound of $I(Z,X)$.

\subsection{Confidence-based Reward Model Ensembling}
It is common that diverse human preferences contain outlier data.
To reduce the influence of a single data point on the reward model, we adopt reward model ensembling to reduce overfitting and improve stability~\cite{pmlr-v139-lee21i,liang2022reward}. Instead of simply averaging, we design a confidence-based ensembling mechanism to improve performance. {\color{black}The key idea is that if the encoder outputs a latent distribution which is far from the reference point (the Gaussian), then the reward model is confident with the data point.} We can aggregate the predicted results by up-weighting the reward models with higher confidence and down-weighting those with less confidence. Specifically, the confidence of a reward model is measured by calculating the KL divergence from the predicted latent space to the pre-defined non-parameterized distribution. If the KL divergence is small which means the reward model cannot tell too much input-specific information, the reward model has low confidence about the input (a model tends to output the non-parameterized distribution directly if it knows nothing about the input). On the contrary, a large KL divergence indicates high confidence.
Formally, the confidence is calculated by:
\begin{equation}
\label{confidence}
    G_i(s,a)=\frac{\exp (\text{KL}(p_i(z|s,a)||r(z)))}{\sum_{j=1}^N \exp (\text{KL}(p_j(z|s,a)||r(z)))},
\end{equation}
where $G_i(s,a)$ denotes the confidence of the $i$-th reward model to an input $(s,a)$, $N$ is the number of reward models.
After determining the confidence of each model, we calculate the reward by:
\begin{equation}
\label{reward}
    \hat{r}(s,a;\psi)=\sum_{i=1}^N G_i(s,a) \times \Bigl[d_i \circ q_i(r|s,a)\Bigr],
\end{equation}
where $d_i$ and $q_i$ are the decoder and the encoder of the $i$-th reward model, respectively. With the reword model, we can use any preference-based RL algorithm to learn the policy. 
The full procedure of our method is summarized in Algorithm~\ref{alg}.

\begin{figure*}
    \centering
\includegraphics[width=1\textwidth]{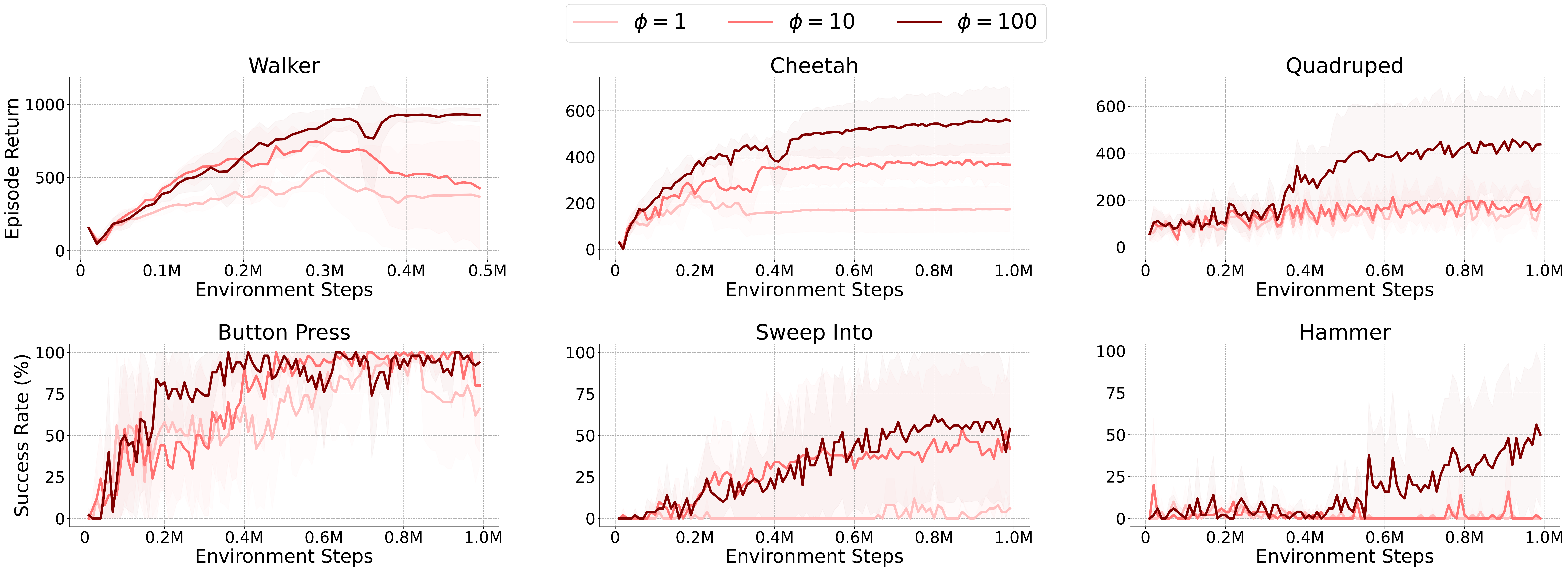}
    \caption{Ablation study on the strength of the latent space constraint. The locomotion tasks (first row) are measured on the ground truth episode return while the robotic manipulation (second row) tasks are measured on the success rate. The results show the mean and standard deviation averaged over five runs.}
    \label{fig:ablation_constrain}
\end{figure*}

\section{Experiments}
{\color{black}
We conduct experiments to answer the following questions:
    \textit{Q1:} How does the proposed method perform under diverse human feedback?
    \textit{Q2:} How does each component of the method contribute to the effectiveness?
    \textit{Q3:} How the predicted rewards are manipulated?
    \textit{Q4:} How will the method be affected by the number of annotators?}

\subsection{Setup}
We evaluate our method on several complex locomotion tasks and robotic manipulation tasks from DeepMind Control Suite (DMControl) \cite{tassa2018deepmind,tunyasuvunakool2020dm_control} and Meta-world~\cite{yu2020meta}, respectively (see Fig.~\ref{fig:demon}). In order to justify the effectiveness of our method, we train an agent to solve the tasks without observing the true rewards from the environment. Instead, several scripted annotators are generated to provide their preferences between two trajectory segments for the agent to learn its policy. Unlike existing preference-based RL algorithms which interact with a single perfect scripted teacher~\cite{christiano2017deep,pmlr-v139-lee21i,park2022surf}, we consider the situation where there is a team of different annotators with bounded rationality to provide the preferences. Despite being imperfect, the annotators' preferences are also calculated from ground truth rewards. Therefore, we can quantitatively evaluate the method by measuring the true episode return or success rate from the environments.

Our method can be integrated into any preference-based RL algorithm to recover their performance under diverse preferences. In our experiments, we choose one of the most popular approaches, PEBBLE~\cite{pmlr-v139-lee21i}, as the backbone algorithm. We examine the performance of PEBBLE under i) a perfect scripted teacher who provides the ground true feedback (oracle); ii) a team of randomly sampled annotators whose feedback is imperfect and anisotropic. The goal of is to recover the performance of PEBBLE under the annotators as much as possible, to approach the oracle case.

\begin{figure*}
    \centering
    \includegraphics[width=0.99\textwidth]{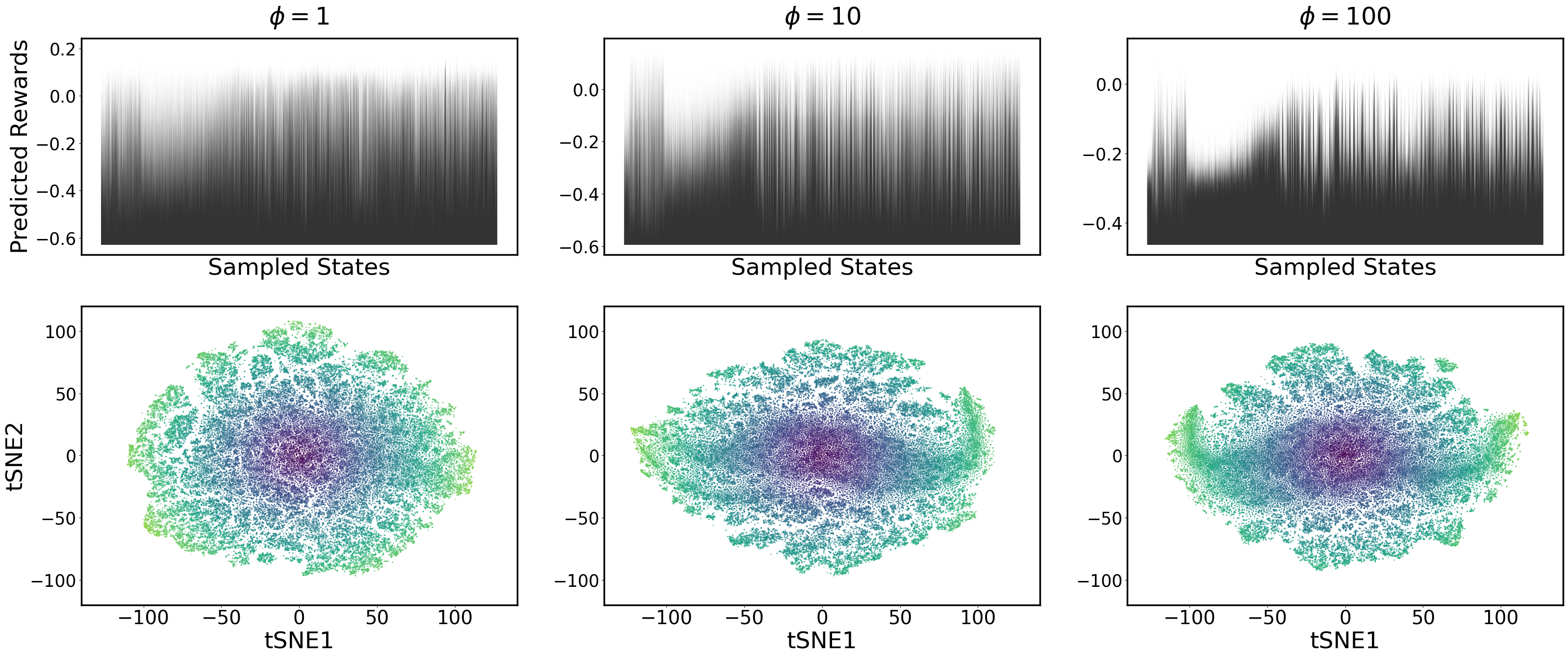}
   \caption{{\color{black}Analysis about the influence of $\phi$ on the reward model. \textbf{First row:} increasing the strength of the constraint will narrow the value range of the predicted rewards. The reward model will also generate more distinct predictions if $\phi$ is large. \textbf{Second row:} the t-SNE visualization of the latent vectors. A large $\phi$ leads to a more compact and concise pattern.}}
    \label{fig:visulization}
\end{figure*}

\textbf{Simulating the annotators.} We generate the bounded rational scripted annotators by following the previous stochastic preference model~\cite{lee2021bpref}:
\begin{equation}
    P\left[\sigma^{1} \succ \sigma^{0}\right]=\frac{\exp \left(\beta\sum_{t=1}^H \gamma^{H-t}{r}\left({s}_{t}^{1}, {a}_{t}^{1} \right)\right)}{\sum_{i \in\{0,1\}} \exp \left(\beta\sum_{t=1}^H \gamma^{H-t}{r}\left({s}_{t}^{i}, {a}_{t}^{i} \right)\right)}.
\end{equation}
$r(s,a)$ is the ground truth reward provided by the environment.
A scripted annotator is determined by a tuple $\langle \beta, \gamma, \epsilon, \delta_\text{equal}\rangle$: $\beta$ is the temperature parameter that controls the randomness of the stochastic preference model. An annotator becomes perfectly rational and deterministic as $\beta \rightarrow \infty$, whereas $\beta=0$ produces uniformly random choices.
$\gamma$ controls the myopic (short-sighted) behavior of an annotator. Annotators with small $\gamma$ will emphasize more on immediate rewards and down-weight long-term return. $\epsilon$ describes the probability that an annotator makes a mistake, i.e., we flip the preference with the probability of $\epsilon$. $\delta_\text{equal}$ denotes the threshold that an annotator marks the segments as equally preferable, i.e., an annotator provides $(0.5,0.5)$ as a response if $|\sum_t r(s_t^1,a_t^1)-\sum_t r(s_t^0,a_t^0)|\leq\delta_\text{equal}$. Practically, we sample a tuple from $\beta \in \{\infty, 1, 5\}$, $\gamma \sim U(0.8,1)$, $\epsilon \sim U(0,0.2)$, $\delta_\text{equal} \sim U(0,0.2)$ to generate a scripted annotator. For each task, we randomly generate $100$ annotators to provide feedback. Each annotator has the same probability of being selected for annotation. 

\textbf{Implementation details.} For all tasks, we use the same hyperparameters used by PEBBLE, such as learning rates, architectures of the neuron networks, and reward model updating frequency. We adopt an uniform sampling strategy, which selects queries with the same probability. At each feedback session, a batch of 256 trajectory segments $(\sigma^0, \sigma^1)$ is sampled for annotation. The strength of constraint $\phi$ is set as 100 for all tasks. For simplicity, we set the reference distribution of the latent space as standard Gaussian where the KL-divergence from a latent space to its prior can be easily calculated. All experimental results are reported with the mean and standard deviation across five runs.

\begin{figure}
    \centering
    \includegraphics[width=1\linewidth]{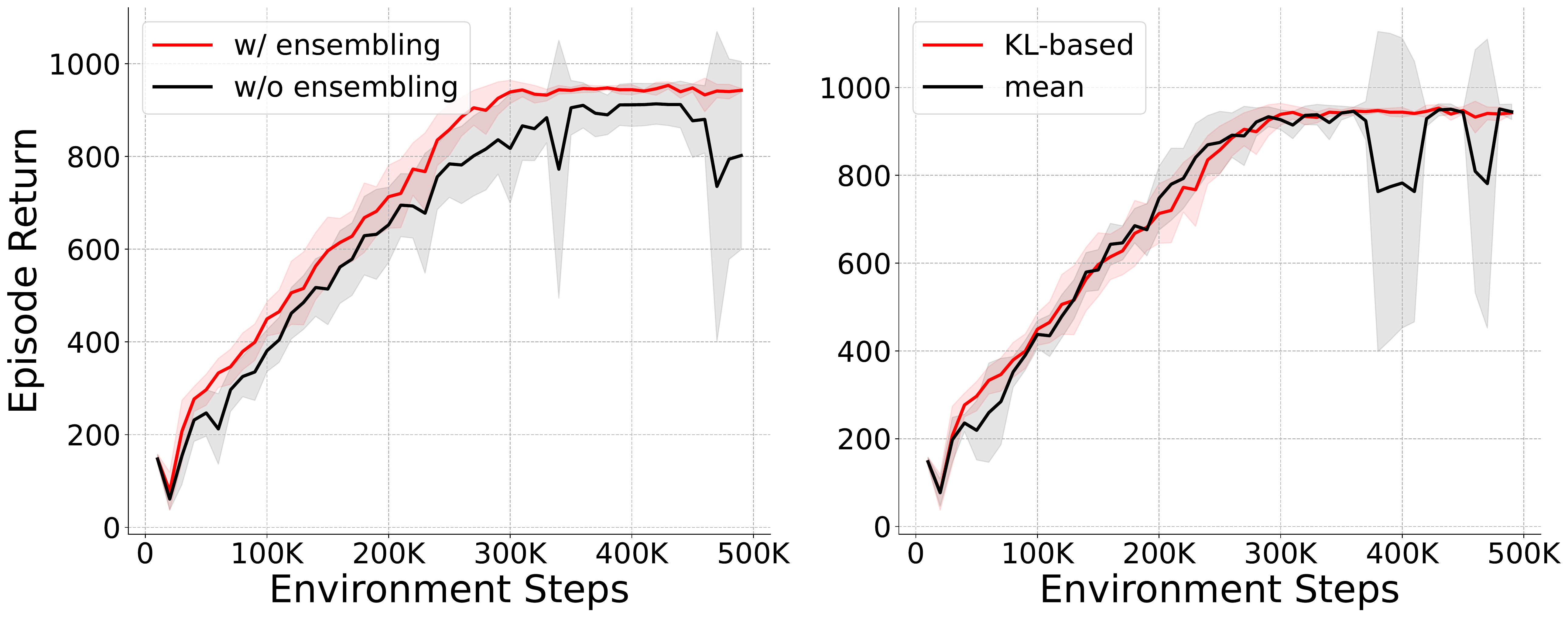}
    \caption{Ablation study on reward model ensembling. \textbf{Left:} Learning curves of Walker\_walk with and without reward model ensembling. \textbf{Right:} Learning curves of Walker\_walk with KL-based model ensembling and simply averaging. The results show the mean and standard deviation averaged over five runs.
}
    \label{fig:ablation_en}
\end{figure}

\subsection{Experimental Results}
\textbf{Locomotion tasks from DMControl.} We select three complex environments from DMControl, which are Walker\_walk, Cheetah\_run, and Quadruped\_walk, to evaluate our method. As previously mentioned, PEBBLE is used as the backbone algorithm and our method is combined with PEBBLE to recover its performance under diverse human preferences. The first row of Fig.~\ref{fig:main} shows the learning curves of PEBBLE and our method when learning from bounded rational annotators. We can find that, compared to learning from a single perfect teacher (oracle), the performance of PEBBLE (measure on true episode return) decreases dramatically in all three tasks. For example, in Walker\_walk, PEBBLE is able to achieve 1000 scores if it learns from a single expert, whereas the value is nearly half of the preferences are from different annotators. Such failures show that existing preference-based RL algorithms do not work well when the provided preferences contain diversity and inconsistency.
After integrating our method (the red line), we can find significant performance increases in all three tasks: our method is able to achieve almost the same performance as the oracle in Walker\_walk, while the performance gap between our method and its upper bound (oracle) is very narrow in Cheetah\_run. In Quadruped\_walk, PEBBLE is unable to learn a feasible policy, while our proposed method still achieves near-optimal performance. 

\textbf{Robotic manipulation tasks from Meta-world.}
Meta-world consists of 50 tasks that cover a range of fundamental robotic manipulation skills. We consider three challenging environments to evaluate the effectiveness of our method. The performance is measured on success rate, i.e., if the trained agent is able to successfully finish a task or not. Fig.~\ref{fig:main} (second row) shows the learning curves of our method as the baselines. We can find that there is a significant performance increase after integrating our method into PEBBLE. The performance can be almost restored to approach the oracles in Button Press and Hammer, while in Sweep Into, the improvement is also non-trivial. These results again demonstrate that our proposed method is able to effectively help the existing preference-based reinforcement learning algorithms learn from diverse preferences.

\begin{figure*}
    \centering
    \includegraphics[width=1\textwidth]{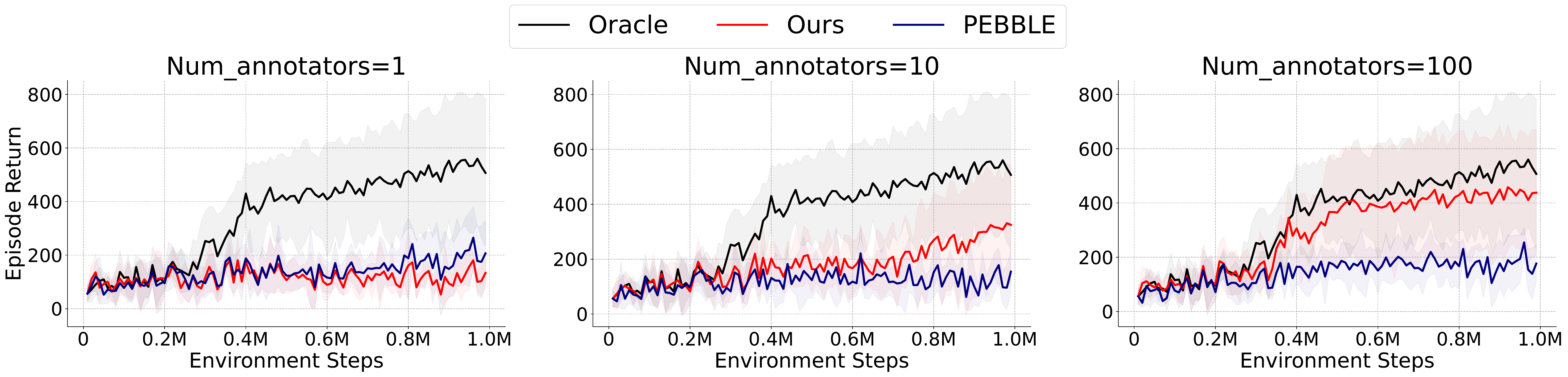}
    \caption{Learning curves of Cheetah\_run with preferences provided by different numbers of annotators. Results are measured on the ground truth episode return, with the solid line and shaded regions representing the mean and standard deviation, respectively, across five runs.}
    \label{fig:ablation_n_annotators}
    \vspace{-0.2cm}
\end{figure*}
\subsection{Ablation and Analysis}
\textbf{Effects of the latent space constraint.}
As previously introduced, to achieve temporal consistency and set a fixed optimization direction for the reward model, our method imposes a constraint on the latent space by forcing its distribution to be close to the prior. To justify the effect of the constraint, we implement our method with different strengths of constraint on all six tasks. As shown in Fig.~\ref{fig:ablation_constrain}, there is a significant and consistent improvement in all the tasks as we gradually increase the strength of the constraint. For example, in Cheetah\_run, when we set the strength of constraint $\psi$ to 1, 10, and 100, the performance increases from around 200 to 400 and finally reaches near 600.
This phenomenon is a little bit counter-intuitive because, conventionally, a too strong constraint is likely to misguide the reward model and prevent it from learning from supervised signals. However, we found that a strong constraint on the latent space is compulsory to help an agent learn from diverse feedback. 

\textbf{Analysis about the reward model.} To understand why a strong constraint is crucial for the performance, we analyze the effect of the latent space constraint on the reward model. Specifically, we collect 100,000 state-action pairs from Cheetah\_run at different training stages and use the reward model trained with different strengths of constraint to predict the rewards. The predictions are presented in the first row of Fig.~\ref{fig:visulization}. We can find that as we increase the strength of constraint, the range of reward values decreases gradually. For example, when $\phi=1$, the predicted rewards are between -0.6 and 0.15, while the range is narrowed to $[-0.5, 0]$ when we set $\phi=100$. Moreover, the predicted rewards are more distinct when $\phi$ becomes larger, which indicates that only really good state-action pairs are given high rewards. We also use t-SNE~\cite{van2008visualizing} to visualize the latent vector of those state-action pairs. As in Fig.~\ref{fig:visulization} (second row), the distribution of the embeddings becomes more compact as we increase $\phi$. Furthermore, the pattern of the embeddings is more clear and more concise when $\phi$ is large. 

\textbf{Effects of reward model ensembling.}
We investigate how well the reward ensembling affects the performance of our method. Fig.~\ref{fig:ablation_en} (left) shows the learning curves of Walker\_walk with and without reward model ensembling. We can find that there is a clear performance drop if using a single reward model. The results demonstrate that ensembling the predictions of several models is helpful to stabilize the training process if the preferences are diverse. We further investigate whether the proposed KL-based aggregation method is better than simply averaging.  As shown in Fig.~\ref{fig:ablation_en} (right), simply averaging will suffer severe fluctuations in training, while our method is significantly more stable and consistent.

\textbf{Responses to the number of annotators.} To examine how will our algorithm respond to the number of annotators, we implement Cheetah\_run with preferences provided by different numbers of annotators. As shown in Fig~\ref{fig:ablation_n_annotators}, when there is only one annotator which is bounded rational, both PEBBLE and our method perform worse than the oracle. In these cases, the preferences are not diverse but just partially correct.
If we slightly increase the number of annotators to ten, which introduces some diversity, our method is able to achieve obvious improvement over PEBBLE. The performance gain becomes quite significant when there are one hundred annotators. The experiments demonstrate that our method is suitable for situations where the provided preferences are diverse.

\section{Related Work}
The main focus in the paper is on one promising direction which utilizes human preferences~\cite{akrour2011preference,christiano2017deep,ibarz2018reward,leike2018scalable,pmlr-v139-lee21i,ouyang2022training,park2022surf,PrefRec,liang2022reward} to perform policy optimization.  Christiano et al. introduced modern deep learning techniques to preference-based learning~\cite{christiano2017deep}. Since the learning of the reward function, modeled by deep neural networks, requires a large number of preferences, recent works have typically focused on improving the feedback efficiency of a method. PEBBLE~\cite{pmlr-v139-lee21i} proposed a novel unsupervised exploration method to pre-train the policy. SURF~\cite{park2022surf} adopted a semi-supervised reward learning framework to leverage a large number of unlabeled samples. Some other works improved data efficiency by introducing additional types of feedback such as demonstrations~\cite{ibarz2018reward} or non-binary rankings~\cite{cao2021weak}. 
In addition to that, designing intrinsic rewards to encourage effective exploration is also investigated~\cite{liang2022reward}. Despite being efficient, previous methods mainly focus on learning from a single expert, which will severely limit the scalability of an algorithm. Recently, there is a line of work on multi-objective RLHF focusing on capturing different intentions in crowds~\cite{zhou2024beyond}. 
In our work, the focus is on addressing the fluctuations in reward learning and policy collapse. It is therefore orthogonal to those multi-objective RLHF methods.

\section{Conclusion}
{\color{black}In this work, we propose a simple yet effective method aimed at enabling RL agents to learn from diverse human preferences. The method stabilizes the reward learning by imposing strong constraint to the latent space of rewards, controlling the divergence between the latent space and a non-parameterized distribution. Furthermore, a confidence-based model ensembling approach is proposed to better aggregate rewards. The effectiveness of the method is demonstrated on a variety of complex locomotion and robotic manipulation tasks. Our method significantly improves over existing preference-based RL algorithms in all tasks when learning from diverse human feedback.}

% \section*{Ethical Statement}

% There are no ethical issues.

\section*{Acknowledgments}
This research is supported by the National Research Foundation, Singapore under its Industry Alignment Fund – Pre-positioning (IAF-PP) Funding Initiative. Any opinions, findings and conclusions or recommendations expressed in this material are those of the author(s) and do not reflect the views of National Research Foundation, Singapore.

\bibliographystyle{named}
\bibliography{ijcai24}

\end{document}